\documentclass[letterpaper]{article} 
\usepackage[draft]{aaai2026}
\usepackage{times}  
\usepackage{helvet}  
\usepackage{courier}  
\usepackage[hyphens]{url}  
\usepackage{graphicx} 
\usepackage{natbib}  
\usepackage{caption} 
\usepackage{algorithm}
\usepackage{algorithmic}

\usepackage{colortbl} 

\usepackage{graphicx}
\usepackage{amsmath, amsfonts}
\usepackage{booktabs} 
\usepackage{pifont}
\usepackage{multirow}
 \usepackage{makecell}
 \usepackage[T1]{fontenc}

\usepackage{newfloat}
\usepackage{listings}
\DeclareCaptionStyle{ruled}{labelfont=normalfont,labelsep=colon,strut=off} 
\floatstyle{ruled}
\newfloat{listing}{tb}{lst}{}
\floatname{listing}{Listing}
\title{UniMapGen: A Generative Framework for Large-Scale Map Construction \\ from Multi-modal Data}

\author{
    Yujian Yuan\textsuperscript{\rm 1,2}\thanks{Work done during internship at Amap, Alibaba Group.}\equalcontrib,
    Changjie Wu\textsuperscript{\rm 1}\equalcontrib,
    Xinyuan Chang\textsuperscript{\rm 1}\equalcontrib,
    Sijin Wang\textsuperscript{\rm 1}\equalcontrib,\\
    Hang Zhang\textsuperscript{\rm 1},
    Shiyi Liang\textsuperscript{\rm 1,3}\footnotemark[1],
    Shuang Zeng\textsuperscript{\rm 1,3}\footnotemark[1],
    Mu Xu\textsuperscript{\rm 1}\thanks{Corresponding author.},
    Ning Guo\textsuperscript{\rm 1}
}
\affiliations{
    \textsuperscript{\rm 1}Amap, Alibaba Group\\
    \textsuperscript{\rm 2}The Hong Kong University of Science and Technology\\
    \textsuperscript{\rm 3}Xi’an Jiaotong University\\
    \tt\small yyuanbn@connect.ust.hk, \{wuchangjie.wcj,changxinyuan.cxy,sijin.wsj,suishou.zh,xumu.xm\}@alibaba-inc.com, \{zengshuang,sy\_liang2023\}@stu.xjtu.edu.cn, ning.guo@alibaba-inc.com

}

\usepackage{bibentry}

\begin{document}

\maketitle

\begin{abstract}
Large-scale map construction plays a vital role in applications like autonomous driving and navigation systems.
Traditional large-scale map construction approaches mainly rely on costly and inefficient special data collection vehicles and labor-intensive annotation processes.
While existing satellite-based methods have demonstrated promising potential in enhancing the efficiency and coverage of map construction, they exhibit two major limitations: 
 (1) inherent drawbacks of satellite data (e.g., occlusions, outdatedness)
and (2) inefficient vectorization from perception-based methods, resulting in discontinuous and rough roads that require extensive post-processing.
This paper presents a novel generative framework, UniMapGen, for large-scale map construction, offering three key innovations: 
(1) representing lane lines as \textbf{discrete sequence} and establishing an iterative strategy to generate more complete and smooth map vectors than traditional perception-based methods.
(2) proposing a flexible architecture that supports \textbf{multi-modal} inputs, enabling dynamic selection among BEV, PV, and text prompt, to overcome the drawbacks of satellite data.
(3) developing a \textbf{state update} strategy for global continuity and consistency of the constructed large-scale map.
UniMapGen achieves state-of-the-art performance on the OpenSatMap dataset. 
Furthermore, UniMapGen can infer occluded roads and predict roads
missing from dataset annotations.
Our code will be released.

\end{abstract}

\begin{figure*}[t]
    \centering
    \includegraphics[width=1.0\textwidth]{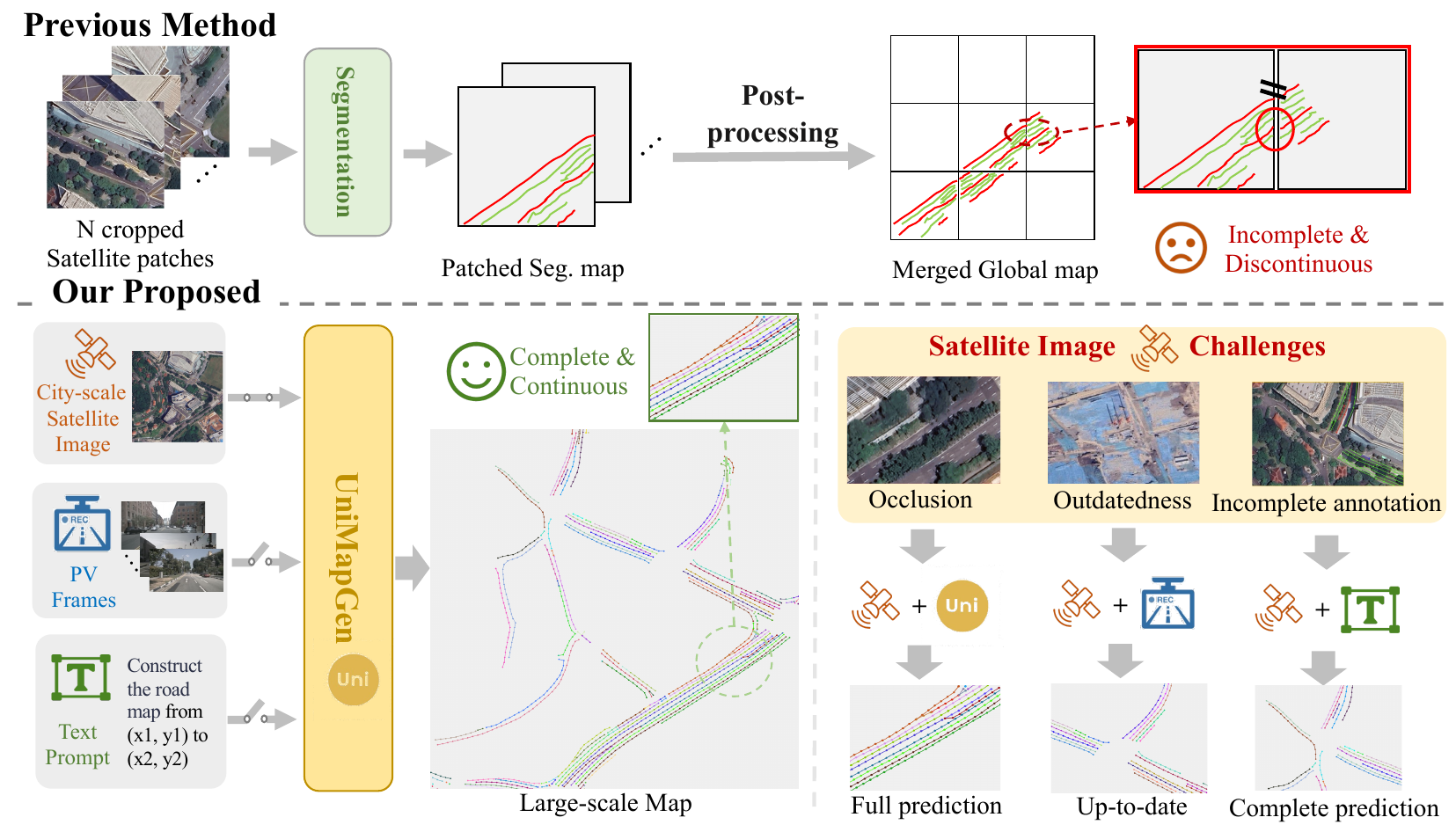}
    \vspace{-1.5 em}
    \caption{
     Methods and challenges in large-scale map construction. (\textbf{Top}) Previous segmentation methods process image patches separately, causing incomplete and discontinuous lines. (\textbf{Bottom}) UniMapGen uses flexible multi-modal inputs to construct complete and continuous maps, overcoming satellite challenges including occlusion, outdateness, and incomplete annotation.}
    \label{fig:figure1}
\end{figure*}


\section{Introduction}

Lane-level map navigation is foundational for critical applications such as autonomous driving and human driving systems.
especially as intelligent transportation applications expand across cities. The construction of large-scale, high-quality vector maps has thus become a critical research challenge, as such maps are essential for enabling precise localization and route planning.
Traditional Standard Definition (SD) maps lack precise lane-level information, resulting in limited navigation capabilities.
While High Definition (HD) maps~\cite{liu2023vectormapnet} offer comprehensive vector and attribute data, their creation demands substantial resources, particularly in terms of specialized data collection vehicles.
The use of satellite image to efficiently construct large-scale high-quality, lane-level maps that fall between SD and HD, as proposed by OpenSatMap~\cite{zhao2024opensatmap}, has become a trend due to its advantage of satisfying global geographical coverage.

When constructing large-scale lane-level maps from satellite image, three essential challenges are inevitable (Fig.~\ref{fig:figure1} right bottom):
(1) Occlusion of roads~\cite{6460256}. 
The physical occlusions (e.g., by tall buildings and trees) in satellite image can cause missing or inaccurate road predictions.
(2) Outdatedness of satellite image. 
The irregular collection of satellite image leads to time gaps between annotations and up-to-date road conditions, which hinders map accuracy.
(3) Incomplete annotations. 
 As the definition of roads varies across different annotators, there are various inconsistent and incomplete road annotations in the datasets.

Existing satellite-based methods for map construction are mainly segmentation-based approaches (e.g., SegNeXt~\cite{zhao2024opensatmap}) and detection-based vectorized methods~\cite{liao2022maptr,liu2023vectormapnet}.
They fail to address the above fundamental challenges inherent in satellite image. Moreover, they overlook critical aspects of electronic map representation, which require vector formatting with global consistency.
These limitations lie in three key areas:
(1) Inefficient representation. 
Existing segmentation-based approaches require complex post-processing, while detection-based methods use fixed-point vectors, 
which is limited for complex long lines and wasteful for short ones.
(2) Discontinuous connecting.
Both segmentation and detection methods require post-processing for connecting together multiple patches, leading to discontinuous connections (Fig.~\ref{fig:figure1} top).
(3) Limited flexibility. Existing frameworks demonstrate limited flexibility as they support fixed modality input (e.g., Bird's Eye View (BEV) only). 
This harms the framework’s ability to improve map construction using alternative inputs, such as fresher Perspective View (PV) frames or specialized road mapping prompts.

To address the above critical issues,
inspired by the powerful contextual learning, inference, and flexible multi-modal fusion ability of multi-modal large language models (MLLMs)~\cite{liu2023visual,li2023blip,chen2024internvl}, we propose UniMapGen, a  MLLM-based novel framework for large-scale map construction. As shown in Fig.~\ref{fig:figure1}, it offers three key innovations:
(1) We reformulate lane-level map construction as a token-based generative problem using a generative model backbone. 
This approach produces smoother vector outputs across lane lines of varying lengths.
(2) We propose a state update strategy for constructing large-scale maps. The map updated at each state relies on that of the last state. 
(3) Our framework accepts flexible multi-modal inputs, including PV, BEV, and text prompt inputs.
PV images reduce inaccurately constructed maps caused by occlusions and outdateness in satellite image. 
Text prompt offers flexibility for interactive map construction,  eliminating the problem of incomplete annotation.
Extensive experiments on OpenSatMap~\cite{zhao2024opensatmap} demonstrate the effectiveness of UniMapGen.

In summary, our contributions include:

\begin{itemize}

    \item We innovatively formulate large-scale map construction as an iterative generation task. 
    To achieve this, we propose to serialize the line vectors and employ a state update strategy that iteratively builds the large-scale maps.
    It improves the map smoothness and completeness by effectively using image and map context.
    
    \item We propose UniMapGen, a novel token-based generative framework that unifies multi-modal inputs for flexible map construction. 
    UniMapGen integrates BEV image, PV frames, and text prompt as input, 
    supporting any combination of them.
    This multi-mode approach addresses the occlusion, outdatedness, and incomplete annotations issues of map construction from satellite image.

    \item UniMapGen demonstrates SOTA performance on OpenSatMap datasets at zoom level-20. 
    Additionally, UniMapGen can infer occluded roads and predict roads missing from dataset annotations.
    The code and trained models will be released soon.

\end{itemize}

\section{Related Work}

\textbf{Large-Scale Map Construction.} 
Traditional large-scale map construction primarily relies on semantic segmentation approaches~\cite{7873262,zhou2018d,demir2018deepglobe,jiang2022roadformer,cheng2025roadnet}. Notable advances include D-LinkNet's~\cite {zhou2018d} multi-scale dilated convolution with pretrained ResNet34~\cite{he2016deep}, RoadFormer's~\cite{jiang2022roadformer} pyramidal deformable vision transformer, and CE-RoadNet's~\cite{cheng2025roadnet} cascaded CNN architecture, each improving road extraction through enhanced feature learning. RoadDA~\cite{zhang2021stagewise} further addresses domain adaptation through adversarial learning and self-training for cross-domain performance.
However, segmentation-based methods face common challenges in maintaining structural continuity and struggle with lane-level map generation. Our work addresses these limitations through a generative end-to-end framework that directly produces vectorized road maps from satellite images, eliminating post-processing artifacts and enabling accurate lane-level map generation.

\textbf{Vertorized Map Construction.}
The evolution of vectorized map construction began with HDMapNet's~\cite{li2022hdmapnet} two-stage segmentation approach, followed by VectorMapNet's~\cite{liu2023vectormapnet} pioneering end-to-end framework using sequential point prediction. MapTR~\cite{liao2022maptr} introduced unified shape modeling with parallel processing, inspiring subsequent developments: MapTRv2~\cite{liao2024maptrv2} enhanced training through auxiliary supervision, PivotNet~\cite{ding2023pivotnet} preserved geometry using dynamic pivotal points, and P-MapNet~\cite{jiang2024p} improved robustness via masked pre-training. Recent advances include MapTracker's~\cite{chen2024maptracker} temporal consistency through query propagation and SMART's~\cite{ye2025smart} integration of map priors from SD Map and satellite image.
However, these detection-based methods rely on fixed-point vector representations regardless of line length, leading to potential quality degradation for very short or long lines.
To address this, we propose a equidistant sampling strategy combined with a generative framework that supports variable-length line prediction, enabling more accurate and adaptive vector representations.

\begin{figure*}[t]
    \centering
    \includegraphics[width=1.0\textwidth]{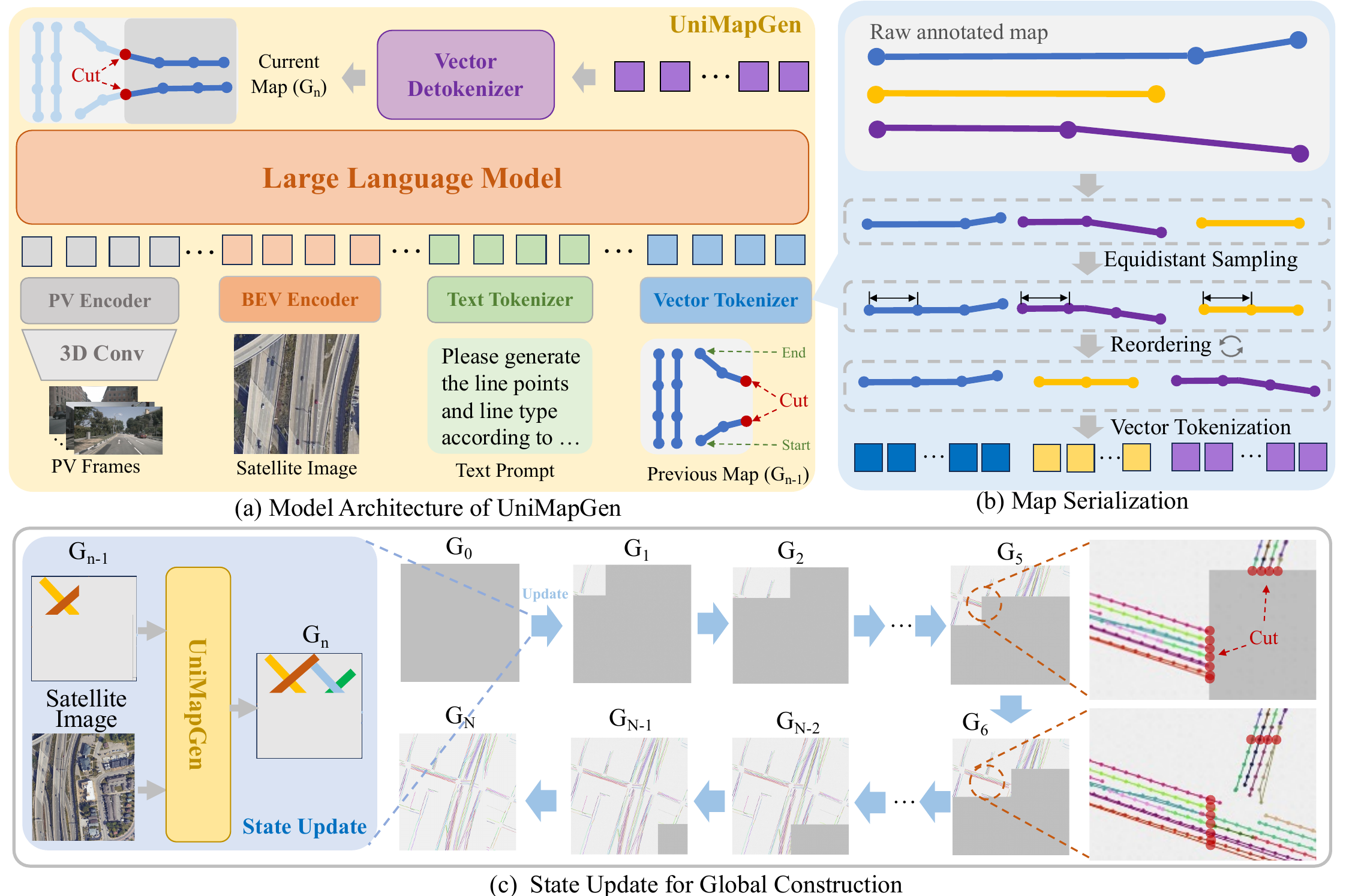}
    \vspace{-2 em}
    \caption{Overview. (a) \textbf{Model Architecture}: UniMapGen supports multi-modal data inputs, including BEV, PV, text, and maps. (b) \textbf{Map Serialization}: we apply equidistant sampling to the raw map vectors, and then reorder them in the specified order. Finally, they are converted into special tokens.  (c) \textbf{State Update}: we propose a state update strategy to incrementally construct large-scale maps. This process requires no post-processing, yielding smooth and connected outputs.}
    \label{fig:overview}
    \vspace{-1.5 em}
\end{figure*}

\section{Method}

This paper aims to construct large-scale lane-level maps using flexible multi-modal data, primarily relying on city-level satellite image.
An ideal map can be represented as a collection of vectors that are complete and smooth, with global continuity and consistency among them~\cite{ovsjanikov2012functional}.
However, previous methods suffer from inefficient representation, discontinuous connecting, limited flexibility, and drawbacks of satellite image (e.g., occlusion) in map construction.
To solve these, we introduce UniMapGen, a generative framework to construct globally continuous and consistent maps from multi-modal data inputs (e.g., BEV, PV frames, and text) in an iterative state update strategy, overcoming challenges like occlusion and data outdateness.

\subsection{Task Definition}
The large-scale map construction can be formalized as an iterative generation process to achieve the global map $G_N$ at the latest state $N$, the iterative process can be formalized as:
\begin{equation}
G_n = \text{UniMapGen}(I^{BEV}, I^{PV}, T^{Prompt}, G_{n-1}), 
\label{eq:task}
\end{equation}
where $G_n$ represents the map at current state $n$, defined as $G = <V,A> = \{(\boldsymbol{v}_i,\boldsymbol{a_i})\}_{i=1}^K$, K is the number of vectors. Here, $V$ denotes a set of vectors and $A$ represents their corresponding attributes. The vectors $V = \{\boldsymbol{v}_i\}_{i=1}^K$ comprise vectorized lines, where each $\boldsymbol{v}_i = \{(x_j,y_j)\}_{j=1}^{N_j}$, and $(x_j,y_j)$ indicates the geographic position of the j-th point of $\boldsymbol{v}_i$. Each attribute $\boldsymbol{a_i} \in A = \{\boldsymbol{a_i}\}_{i=1}^K$ corresponds to vector $\boldsymbol{v}_i$, encompassing properties such as line categories.

The task accepts multiple optional inputs:
\begin{enumerate}
    \item Bird's-Eye View image ($I^{BEV}\in \mathbb{R}^{H^{BEV}\times W^{BEV} \times 3}$): City-level low-cost satellite image with broad coverage. 
    
    \item Perspective-View frames ($I^{PV} \in \mathbb{R}^{H^{PV} \times W^{PV} \times 3 \times L}$): Ground-level frames that complement satellite data by addressing occlusion and outdated information issues, where L denotes the number of PV frames.
    
    \item Text prompt ($T^{Prompt}$): User-specified task prompts that guide the model to generate specific map regions, enabling interactive road generation and updates.
    
    \item Previous-state map ($G_{n-1}$): The map of the last state $n-1$ that provides context and ensures global continuity and consistency without the need for post-processing.
\end{enumerate}

\subsection{Framework of UniMapGen}\label{sec:framwork}

Existing vector-based map construction methods are perception-based (e.g., detection and segmentation), facing limitations like occlusion and poor vector smoothness.
To address this, we propose to construct the maps using generative large language model (LLM), 
leveraging its powerful inference ability and modality flexibility to generate complete, smooth, and globally continuous vectors.
To enable the generative model to process the map effectively, we first serialize the map representation. Then, to handle multi-modal inputs, we design a multi-modal framework, incorporating specialized encoders and tokenizers for each modality.

\subsubsection{Map Serialization}
Since generative LLMs only support serialized data formats, as shown in Fig.~\ref{fig:overview}(b), we serialize map vectors into a single sequence by: 
(1) resampling original vector representations, and (2) concatenating these discrete vectors in a specified order. 
Traditional fixed-point sampling strategies, such as the 20-point sampling used in MapTR~\cite{liao2022maptr}, struggle to accurately represent long and complex lines in satellite images due to limited point num, while being inefficient for short lines (e.g., 2 m), where many points are redundant. 
To improve this, we propose an equal-distance sampling strategy that samples points at fixed distances, enabling efficient adaptive point allocation based on line length.
In concatenation, we reorder map vectors for order consistency, facilitating easier learning by the LLM, and potentially simplifying map construction.

Formally, given the raw annotated map $G^{r} = \{ (\boldsymbol{v}_i^{r}, \boldsymbol{a}_i) \}_{i=1}^K$,  we perform equidistant sampling on each line vector $\boldsymbol{v}_i^{r}$ at intervals of $N_s$ (we set as 6 meter) to obtain $\boldsymbol{v}_i^{s}$, resulting in equal-distance sampled-map $G^{s} = \{ (\boldsymbol{v}_i^{s}, \boldsymbol{a}_i) \}_{i=1}^K$.
Then, to ensure order consistency, we reorder the vectors $\{ \boldsymbol{v}_i^{s} \}_{i=1}^K$ according to their euclidean distance $d_i$ between their first point to the origin point (i.e., (0,0)), getting $G^{o} = \{ (\boldsymbol{v}_i^{o}, \boldsymbol{a}_i) \}_{i=1}^K, \text{ where } d_{1} \leq d_{2} \leq \dots \leq d_{K}$.
Later, we serialize the map $G^{o}$ to get sequence $S_G$ by concatenating each lane line vector-attribute pairs in order:
\begin{equation}
S_{G} = \mathop{\oplus}_{i=1}^{K} (\boldsymbol{v}_i^o, \boldsymbol{a}_i)
\label{eq:sequence}
\end{equation}

Finally, we convert each part of $S_{G}$ into predefined special tokens. 
Please see \textit{suppl.} for implementation details.

\subsubsection{Model Architecture}
Previous methods receive the fixed-modality input and construct maps with a single mode, limiting the flexibility and efficiency for large-scale map construction, 
as the road information can not be prompted or updated by the inputs from other modalities (e.g., PV frames). 
To address this, as shown in Fig.~\ref{fig:overview}(a), we design an MLLM-based architecture for UniMapGen, leveraging its acceptance of various input modalities and powerful contextual learning ability.
Additionally, to transform the inputs in Eq.\ref{eq:task} into LLM-compatible feature spaces, we designed specialized encoders and tokenizers for each modality.

\begin{table}[!htbp]
    \centering
    \renewcommand{\arraystretch}{2}
    \setlength{\tabcolsep}{0.9pt}
    \footnotesize
    \begin{tabular}{l|c}
    \toprule[2pt]
     & Format \\
    \midrule[1pt]
    Vector & \{'points': [[257,49],...,[376,15]],'category':'Curb',...\} \\
    \midrule[1pt]
    Tokens & \makecell[cl]{<\{>\textbf{<points>}<:><[><[>\textbf{<257>}<,>\textbf{<49>}<]>,...,<[>\textbf{<376>}\\<,>\textbf{<15>}<]><]><,>\textbf{<category>}<:>\textbf{<Curb>},...<\}>}\\
    \bottomrule[2pt]
    \end{tabular}
    \caption{An example of the text format before (vector) and after (tokens) the specialized vector tokenizer. The coordinate number or words are converted into tokens as a whole.}
    \label{tab:tokenizer}
\end{table}

\begin{figure*}[t]
    \centering
    \includegraphics[width=0.6\textwidth]{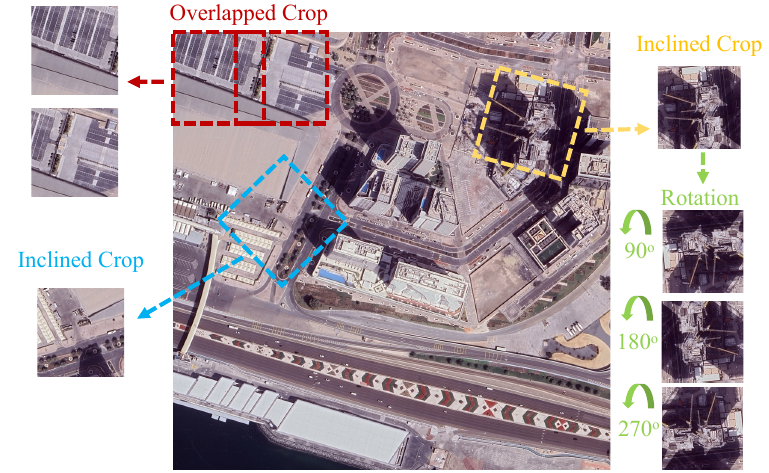}
    \caption{Examples of data augmentation, including overlapped crop and inclined crop with rotation.
    }
    \vspace{1 em}
    \label{fig:examples}
\end{figure*}

\textit{A. Encoders/LLM and Tokenizers.} 
(1) BEV Encoder. We employ ViT-Large-14 from Dinov2~\cite{remoteclip} as our pretrained BEV encoder, encoding a single satellite image $I^{BEV}$ into N D-dimensional tokens $\boldsymbol{e}^{BEV} \in \mathbb{R}^{D \times N}$.
(2) PV Encoder. We utilize a 3D convolution~\cite{tran2015learning} followed by the image encoder from Qwen2-VL-2B~\cite{wang2024qwen2vlenhancingvisionlanguagemodels} as our PV Encoder, processing L PV frames $I^{PV}$ into M D-dimensional tokens $\boldsymbol{e}^{PV} \in \mathbb{R}^{D \times M}$.
(3) Vector Tokenizer and Detokenizer. We propose a special vector tokenizer to convert the previous state map $G_{n-1}$ into special tokens $\boldsymbol{e}^{G}$, subsequently processed through word embeddings. 
The vector detokenizer performs the inverse operation, decoding the MLLM-predicted tokens back into vectorized maps.
Tab.~\ref{tab:tokenizer} illustrates an example.
Specifically, each coordinate number is encoded into special tokens as a whole rather than split into single digits. For example, the number 257 will be tokenized into <257> rather than <2><5><7>. Additionally, the word will be tokenized as a whole as well, such as converting 'points' to <points>. Finally, we remove all the spaces in the token string to reduce the number of tokens.
On the contrary, the vector detokenizer performs the inverse operation, decoding the MLLM-predicted tokens (like the second line) back into vectorized maps (like the first line).
(4) Large Language Model and Text Encoder. We adopt Qwen2.5-1.5B~\cite{qwen2025qwen25technicalreport} as our LLM, receiving multimodal input to construct maps. Additionally, we utilize the LLM's text tokenizer to tokenize $T^{prompt}$ into $\boldsymbol{e}^{T}$.
More model details are listed in \textit{suppl.}

\textit{B. Flexible Multi-modal inputs.} 
To enable flexible map construction, UnimapGen supports multiple generation modes.
The model inputs (BEV image, PV frames, text prompt, and previous map) are designed to be optional, allowing inference with any combination of inputs. The input embedding sequence $S_{input}$ is formulated as:
\vspace{-0.5 em}
\begin{equation}
S_{input} = \boldsymbol{e}^{BEV} \, \| \, \boldsymbol{e}^{PV} \, \| \, \boldsymbol{e}^{T} \, \| \, \boldsymbol{e}^{G},
\label{eq:combined}
\end{equation}
During training, we randomly mask different modalities to enhance sample diversity and develop model's adaptive multi-modal ability. This ensures robust performance regardless of available input modalities.
We use different text prompts to control the generation mode, as listed in \textit{suppl.}

\begin{table*}[t]
    \centering
    \caption{\textbf{Evaluation on OpenSatMap20 validation set}.  AP$^M$ means that the mask IoU is used when determining true positives, while AP$^C_D$ means Chamfer AP with a threshold of $ D $ meters. AP$_x $ denotes that threshold is set to $ x $. AP$ $ indicates averaged values, varying thresholds from 50 to 95. The models noted $^{sat}$ are produced by us using satellite images.}
    \vspace{-1 em}
    \setlength{\tabcolsep}{7pt}
    \footnotesize
    \renewcommand{\arraystretch}{1.2}
    \begin{tabular}{l|c|cccc|cccc}
    \toprule[2pt]
    Models & Method & $\text{AP}^{\text{C}}_{0.9}$ & $\text{AP}^{\text{C}}_{1.5}$ & $\text{AP}^{\text{C}}_{3.0}$ & $\text{AP}^{\text{C}}_{4.5}$ & $\text{AP}^{\text{M}}$ & $\text{AP}^{\text{M}}_{50}$ & $\text{AP}^{\text{M}}_{75}$ & mIoU \\
    \midrule[1pt]
    MapTR$^{sat}$~\cite{liao2022maptr} & vector-based & 18.20 & 22.13 & 26.36 & 28.25 & 6.02 & 14.12 & 4.18 & 34.10 \\
    MapTRv2$^{sat}$~\cite{liao2024maptrv2} & vector-based & 19.34 & 23.44 & 26.98 & 29.89 & 6.45 & 14.89 & 4.47 & 35.42 \\
    \cmidrule(){1-10}
    SegNeXt~\cite{zhao2024opensatmap} & segmentation-based & 20.30 & 25.93 & 29.50 & 31.38 & 6.98 & 16.05 & \textbf{5.26} & 33.69 \\
    \rowcolor{gray!20}
    UniMapGen & vector-based & \textbf{29.17} & \textbf{30.41} & \textbf{32.54} & \textbf{34.81} & \textbf{8.38} & \textbf{21.67} & 5.19 & \textbf{41.81} \\
    \bottomrule[2pt]
    \end{tabular}
    \label{tab:sota result}
    \vspace{-1 em}
\end{table*}

\begin{table}[t]
    \centering
    \caption{\textbf{Ablation on OpenSatMap20 validation set}. Upd means the state update strategy for map construction. Reo means reordering lines, and Aug means data augmentation.
    }
    \vspace{-1 em}
    \footnotesize
    \setlength{\tabcolsep}{1pt}
    \renewcommand{\arraystretch}{1.2}
    \begin{tabular}{ccc|cccc|cccc}
    \toprule[2pt]
     Upd & Reo & Aug & $\text{AP}^{\text{C}}_{0.9}$ & $\text{AP}^{\text{C}}_{1.5}$ & $\text{AP}^{\text{C}}_{3.0}$ & $\text{AP}^{\text{C}}_{4.5}$ & $\text{AP}^{\text{M}}$ & $\text{AP}^{\text{M}}_{50}$ & $\text{AP}^{\text{M}}_{75}$ & mIoU \\
    \midrule[1pt]
     \ding{55}  & \ding{55} & \ding{55} & 21.44 & 22.68 & 24.76 & 26.05 & 4.92 & 14.44 & 2.28 & 36.84 \\
     \ding{55}  & \ding{55} & \ding{51} & 24.53 & 25.82 & 28.19 & 29.57 & 6.40 & 17.32 & 3.57 & 38.99 \\
     \ding{51}  & \ding{55} & \ding{51} & 25.89 & 26.96 & 29.78 & 31.29 & 6.92 & 18.42 & 3.99 & 39.37 \\
     \ding{55}  & \ding{51} & \ding{51} & 27.96 & 29.31 & 31.62 & 33.07 & 7.63 & 19.99 & 4.98 & 40.08 \\
    \rowcolor{gray!20}
     \ding{51}  & \ding{51} & \ding{51} & \textbf{29.17} & \textbf{30.41} & \textbf{32.54} & \textbf{34.81} & \textbf{8.38} & \textbf{21.67} & \textbf{5.19} & \textbf{41.81} \\
    \bottomrule[2pt]
    \end{tabular}
    \label{tab:ablation}
    \vspace{-1 em}
\end{table}

\begin{table}[t]
    \centering
    \caption{\textbf{Ablation on OpenSatMap20 val subset on nuSences and Argoverse2 cities}. Target/Full GT: target or full map construction. T: text prompt for target map.}
    \vspace{-1 em}
    \setlength{\tabcolsep}{0.6pt}
    \footnotesize
    \renewcommand{\arraystretch}{1.2}
    \begin{tabular}{l|c|cccc|cccc}
    \toprule[2pt]
    GT&Modal& $\text{AP}^{\text{C}}_{0.9}$ & $\text{AP}^{\text{C}}_{1.5}$ & $\text{AP}^{\text{C}}_{3.0}$ & $\text{AP}^{\text{C}}_{4.5}$ & $\text{AP}^{\text{M}}$ & $\text{AP}^{\text{M}}_{50}$ & $\text{AP}^{\text{M}}_{75}$ & mIoU\\
    \midrule[1pt]
    \multirow{3}{*}{Tar.} & BEV & 9.93 & 11.39 & 13.59 & 16.38& 2.78 & 7.62 & 1.51 & 16.89 \\ 
     & BEV+T & 22.02 & 23.73 & 27.09 & 30.01 & 4.47 & 12.07 & 3.74 & 35.56 \\
    \cellcolor{white} & BEV+T+PV & \textbf{23.28} & \textbf{26.17} & \textbf{30.67} & \textbf{33.59} & \textbf{5.98} & \textbf{14.65} & \textbf{4.26} & \textbf{41.02} \\
    \midrule[1pt]
    \multirow{2}{*}{\cellcolor{white}Full} & BEV  & 29.80 & 31.41 & 33.93 & 35.59 & 8.32 & 22.17 & 4.90 & 43.08  \\
     & BEV+PV  & \textbf{31.91} & \textbf{33.35} & \textbf{35.79} & \textbf{37.11} & \textbf{9.05} & \textbf{23.99} & \textbf{5.43} & \textbf{45.16}\\
    \bottomrule[2pt]
    \end{tabular}
    \label{tab:ablation 2}
    \vspace{-1 em}
\end{table}

\begin{table}[t]
    \centering
    \caption{\textbf{Evaluation on nuSences default split for lane topology construction}.  This evaluates Landmark Precision-Recall, Reachability Precision-Recall, and their F1 score. }
    \vspace{-1 em}
    \setlength{\tabcolsep}{2pt}
    \footnotesize
    \renewcommand{\arraystretch}{1.2}
    \begin{tabular}{l|ccc|ccc}
    \toprule[2pt]
  \multirow{2}{*}{Models} & \multicolumn{3}{c|}{Landmark}                                         & \multicolumn{3}{c}{Reachability}                                      \\
   & L-P           & L-R           & \cellcolor[HTML]{EFEFEF}L-F1           & R-P           & R-R           & \cellcolor[HTML]{EFEFEF}R-F1           \\ \midrule
TopoNet~\cite{li2023graph}             & 52.5          & 47.1          & \cellcolor[HTML]{EFEFEF}49.6          & 46.9          & 10.8          & \cellcolor[HTML]{EFEFEF}17.5          \\
LaneGAP~\cite{liao2024lane}             & 49.9          & 57.0          & \cellcolor[HTML]{EFEFEF}\textbf{53.2 }         & 74.1          & 34.9          & \cellcolor[HTML]{EFEFEF}47.5          \\
RNTR~\cite{lu2023translating}           & 57.1          & 42.7          & \cellcolor[HTML]{EFEFEF}48.9          & 63.7          & 45.2          & \cellcolor[HTML]{EFEFEF}52.8          \\
UniMapGen & 53.4 & 48.7 & \cellcolor[HTML]{EFEFEF}50.9 & 72.6 & 58.3 & \cellcolor[HTML]{EFEFEF}\textbf{64.6} \\ 
    \bottomrule[2pt]
    \end{tabular}
    \label{tab:topo result}
    \vspace{-1 em}
\end{table}

\subsection{State Update for Global Construction}
Previous methods construct large-scale maps by dividing satellite images into patches, processing each independently, and merging results. However, this ignores context from neighboring patches and creates unsmooth lines at patch edges.
To address this, as shown in Fig.~\ref{fig:overview}(c), we employ a state update framework for end-to-end large-scale map construction.
It iteratively builds current map upon the previous map 
while maintaining global continuity and consistency.

To support state update, we add two attributes: start/end type, into $\boldsymbol{a_i}$ of each line vector $\boldsymbol{v_i}$. 
Start/end type refers to the start/end point type of a line, classified as 'start', 'end', or 'cut'. The three types indicate whether the point is a natural start/end point or formed by cutting a line into two patches. 

During the inference process, the state update process is similar to a growth process, where the current map evolves from the previous state map.
Specifically, we adopted a left-to-right, top-to-bottom patch update strategy. 
When constructing the map for a patch, the model refers to all the 'cut' points adjacent to the patch, which are extracted from the previous state map. 
This ensures the edge vector of the next region is consistent with that of the current region.
In this way, the model generates coherent and smooth lane line vectors that seamlessly connect with existing vectors.

Formally, starting from an empty map $G_0$ in initial state, we construct the maps $G_n$ at $n$ state based the maps on $n-1$ state ($G_{n-1}$) and multi-model inputs at $n$ state as:
\begin{equation}
G_{n} = \text{UniMapGen}(\boldsymbol{e}^{BEV}_n \, \| \, \boldsymbol{e}^{PV}_n \, \| \, \boldsymbol{e}^{T}_n \, \| \, \boldsymbol{e}^{G}_{n-1}),
\label{eq:combined}
\end{equation}
After $N$ step iteration, where $N=\frac{H^{BEV}\times W^{BEV}}{H_{patch}\times W_{patch}}$, we obtain the final result $G_N$ with full maps on BEV image $I^{BEV}$.

For the training process of state update, the start/end type of lines comes from the ground truth of the current patch, which is different from that (from adjacent patches) in the inference process.

\section{Experiments}

\textbf{Dataset.} 
We evaluate the performance of UniMapGen on OpenSatMap dataset at 20-level, which includes 1180 training and 393 validation satellite image of 4096*4096 pixels. 
Each image in the dataset is annotated with lane-level vectorized maps, with each line classified into three line categories (Curb, Lane line, and Virtual Line), and eight attributes (e.g., line types and level of occlusion).
To meet the large training data requirement for MLLMs, we conduct overlapped and inclined cropping augmentation, resulting in 700k training satellite patches. Details of the training data construction are listed in the \textit{suppl.}.

\textbf{Data Augmentation.} As the training images of OpenSatMap are quite limited (1180 at 4096*4096). If they are cropped to 896*896 resolution without augmentation, there are only 18880 patches, which is hard to meet the large-scale training data request of Multi-modal Large Language Model. 
The risk of overfitting the MLLM is heightened.
To mitigate this, as shown in Fig.~\ref{fig:examples}, we apply an overlapped and inclined cropping data augmentation to increase the number of training patches from 2K to 700K. 
The i,j-th cropped patch is formatted as:
\begin{equation}
patch_{i,j} = \text{Crop}(I^{BEV}, C_{i,j}, \theta),
\label{eq:distance}
\end{equation}
where the patch is cropped given the satellite image $I^{BEV}$, center point $C_{i,j}$ in the axis of full image of the cropped patch, and the inclined angle $\theta$.
The crop is conducted with stride 664 ($C_{i,j}$ starts from (448, 448)) or 544 ($C_{i,j}$ starts from (1268, 1268)), with the inclined angle of 0, 15, 30, 45, 60, 75 degrees. 
If the cropped patch extends beyond the boundaries of the original satellite image, we discard this patch.
After all the patches are cropped, each patch is rotated by 90, 180, 270 degrees to increase the training samples.

\textbf{Training Data Preparation.} 
To support multi-mode training, beyond BEV-only mode, we prepare other training data across different modalities as follows: (1) PV-only training. We leverage the PV frames from nuSences~\cite{caesar2020nuscenes} and Argoverse2~\cite{wilson2023argoverse} datasets, enabling map construction from PV frames.
To ensure the consistency of training data, we use the modified annotation in OpenSatMap dataset. To be specific, we first use the GPS information provided by nuSences, Argoverse2, and OpenSatMap dataset to link the PV frames to the corresponding BEV images and its annotation. Then, similar to MapTR, we crop the lines within 60m height and 30m width as ground truth, considering the perception field of PV frames.
(2) PV and BEV joint training. 
Similarly, GPS information is also used to link PV frames to BEV images between nuSences/Argoverse2 and OpenSatMap dataset. 
The PV position coordinates in the BEV image and the angle representing the vehicle direction are also incorporated into the input to facilitate more accurate alignment between PV and BEV images.
(3) Text-prompted target map generation. 
We use a semi-automatic annotation pipeline: automatic data synthesis via Segment Anything Model (SAM)~\cite{kirillov2023segment}, followed by human verification and correction for accurate target lane lines and text prompts matching.
Additionally, we also incorporate the trace points from nuSences and Argoverse2 as text prompt for target map generation.
More data preparation details and text prompts are listed in \textit{suppl.}

\textbf{Metric.} Following~\cite{zhao2024opensatmap}, the line width is set to 6 pixels by default, and evaluations are conducted at both semantic and instance levels. 
For semantic-level evaluation, we use the mean intersection over union (mIoU)~\cite{everingham2010pascal} as the metric across different line categories.
For instance-level evaluation, we employ average precision (AP) metrics, which include Mask AP ($\text{AP}^\text{M}$) using mask IoU thresholds to determine true positive samples~\cite{cheng2022masked}, and Chamfer AP ($\text{AP}^\text{C}_\text{D}$)~\cite{zhao2024opensatmap} by choosing the distance threshold D.
Since standard AP computation requires class probabilities for each predicted line, we propose a pseudo-scoring mechanism based on the MLLM's token prediction confidence. Specifically, for the i-th line, we define its pseudo-score $p_i$ as:
\begin{equation}
p_i = \max(\text{softmax}(\mathbf{o}_i))
\label{eq:score}
\end{equation}
where $\mathbf{o}_i \in \mathbb{R}^C$ represents the logits output of MLLM for the i-th line's class token, and C is the number of classes.

\begin{figure*}[t]
    \centering
    \includegraphics[width=0.98\textwidth]{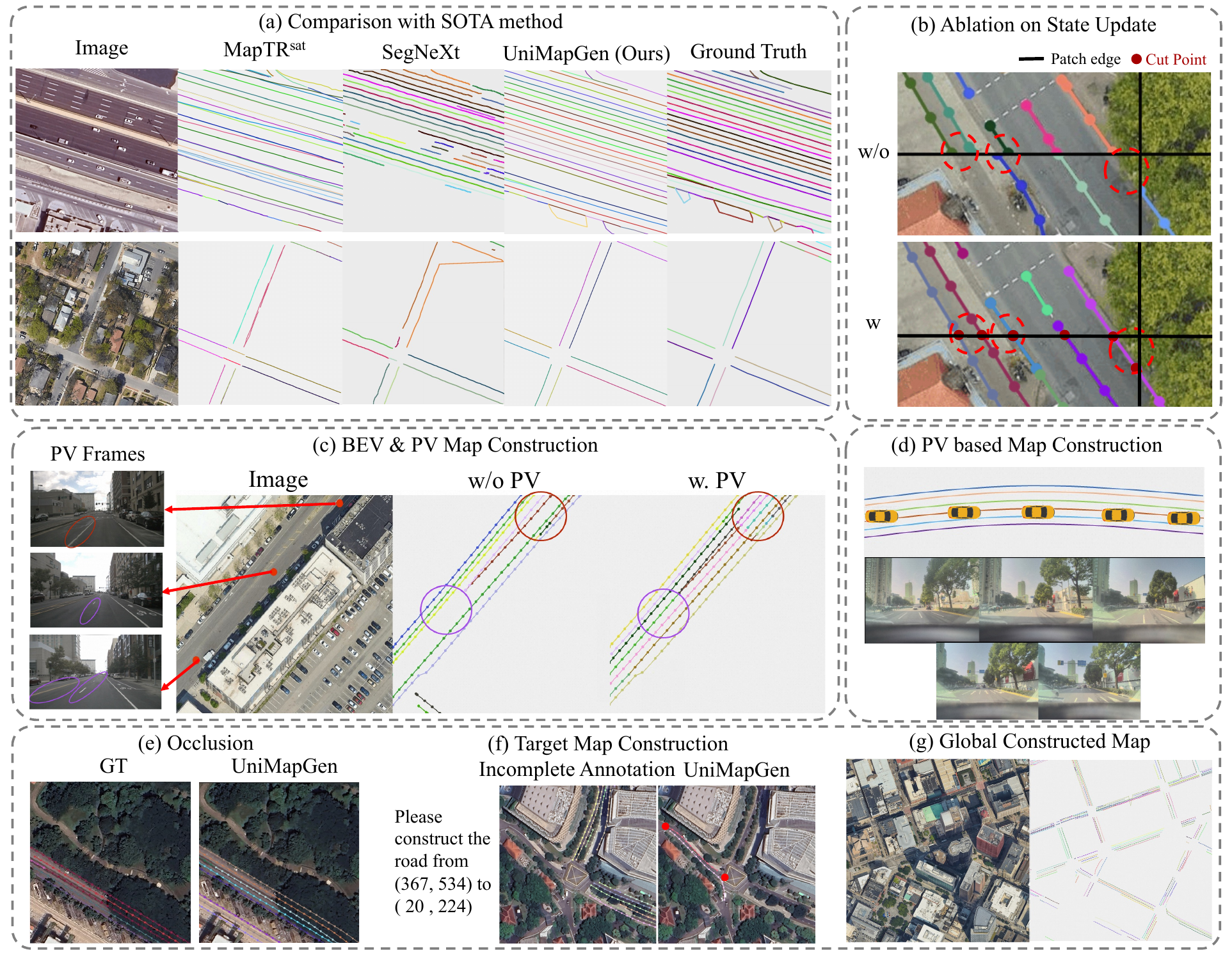}
    \caption{Qualitative results of UniMapGen. (a) Comparison with SOTA. Different color refers to different line instances. (b) Ablation on State Update. The black lines are the patch edges. (c) BEV and PV map construction. PV provides up-to-date (purple) and complementary (red) road information. The line with purple circles is worn out or outdated in BEV image but clear in PV frames. (d) PV-based map construction. (e) Occluded road construction even without PV frames. (f) UniMapGen generates target maps given text prompts. (g) Global constructed map (missing intersection due to OpenSatMap annotation).
    }
    \label{fig:examples}
\end{figure*}

\begin{figure*}[t]
    \centering
    \includegraphics[width=1.0\textwidth]{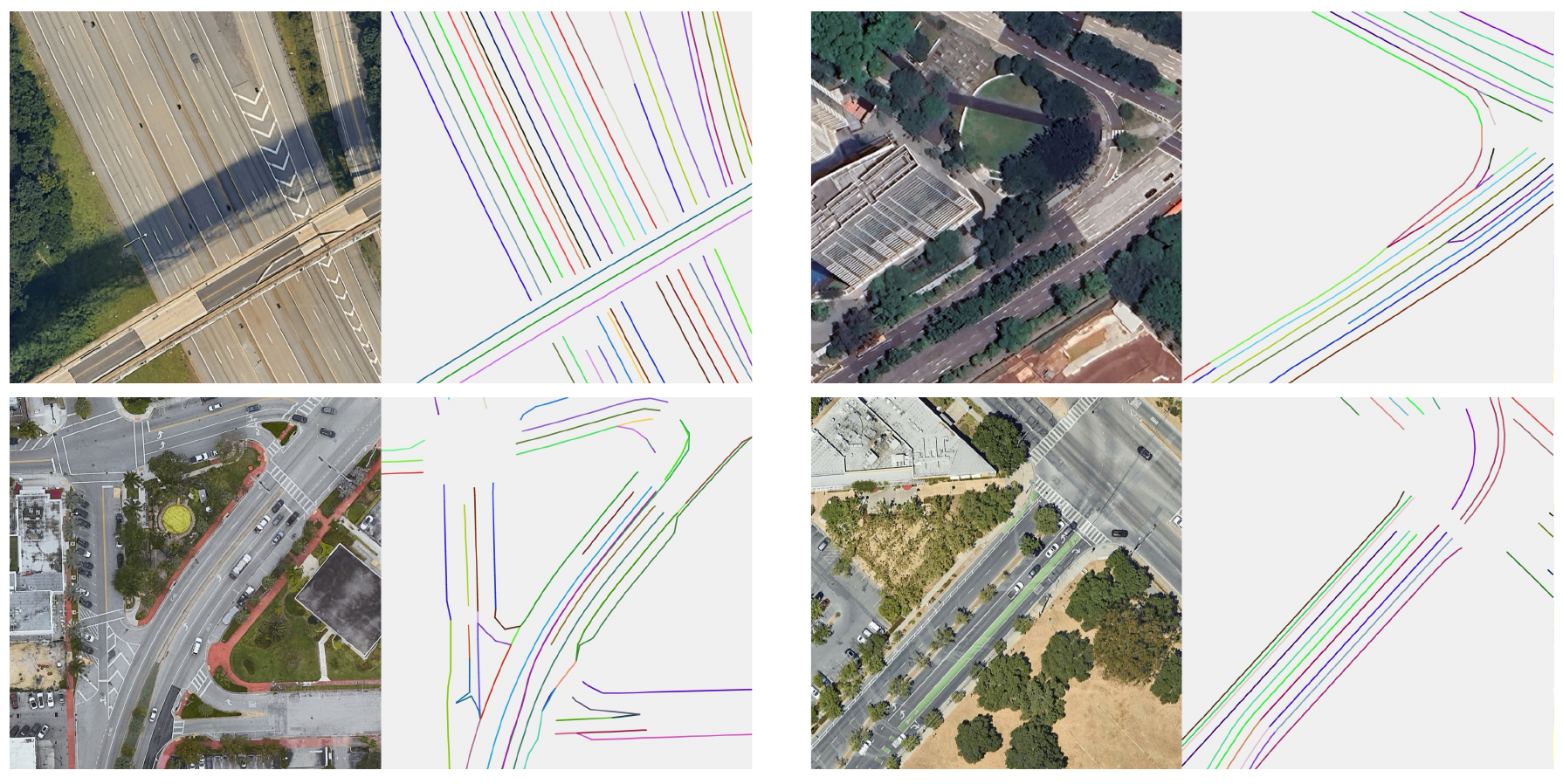}
    \caption{Complex examples generated by UniMapGen. The samples come from different cities.
    }
    \label{fig:complex}
\end{figure*}

\begin{figure*}[t]
    \centering
    \includegraphics[width=1.0\textwidth]{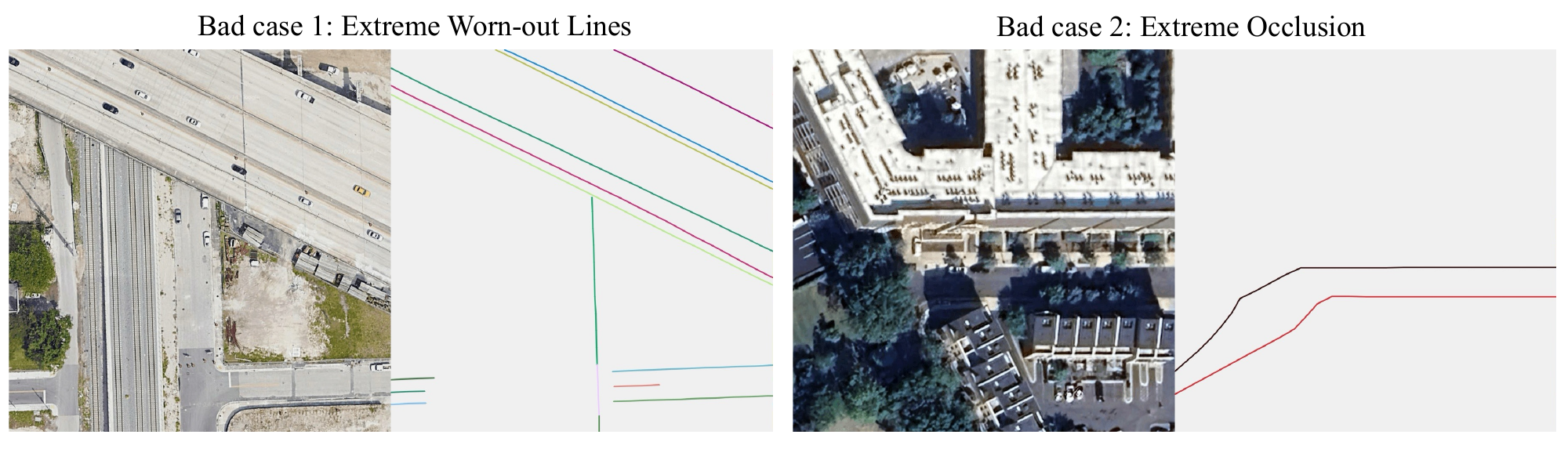}
    \caption{Bad Cases of UniMapGen. Extreme worn-out lines confuse UniMapGen of line's existence. Additionally, extreme occlusion of roads (left bottom) without context also confuses UniMapGen for accurate line location prediction.
    }
    \label{fig:bad case main}
\end{figure*}

\textbf{Implementation Details.} 
For the training of UniMapGen, a batch size 32, and AdamW optimizer with a weight decay of 0.1 are used to fully finetune for 6 epochs. A cosine learning rate decay with a peak learning rate of 2e-5 and a linear warm-up of 100 steps was adopted.
The input satellite image of each state was of size 896×896.
Due to the GPU memory limit, we only use the front-view frames in our experiment. Additionally, we uniformly sample up to 10 PV images for a paired satellite image, and resize each PV image into 644×364.
All the experiments run on 32 Nvidia H20, while UniMapGen can infer on a single Nvidia 3090.

\subsection{Quantitative Evaluation}
We compare UniMapGen with the current state-of-the-art method SegNeXt~\cite{zhao2024opensatmap} and vector-based map construction methods (e.g., MapTR) on the OpenSatMap20 dataset. SegNeXt employs a three-stage pipeline, including semantic segmentation and two post-processing processes (instance detection and vectorization).
It constructs maps for 4096×4096 images by dividing them into 1024×1024 patches and predicting lines for each patch.
MapTR~\cite{liao2022maptr} and MapTRv2 are originally PV frames-based map construction methods. To evaluate their performance of BEV-based generation, 
we replace the BEVFormer with ViT-Large-14 from Dinov2~\cite{oquab2023dinov2}, followed by an MLP, to directly extract BEV features from the satellite image.
Tab.~\ref{tab:sota result} shows that UniMapGen outperforms SegNeXt and MapTR series across most metrics, including mIoU and APs, demonstrating the powerful ability of UniMapGen in large-scale map construction.
Qualitative comparisons in Fig.~\ref{fig:examples} (a) further reveal that UniMapGen generates smoother, more continuous, and complete lines compared to SegNeXt's post-processing and MapTR series results. Fig.~\ref{fig:examples} (g) shows an example of our constructed global map. Additionally, Fig.~\ref{fig:complex} shows several complex examples generated by UniMapGen from 4 different cities.

\subsection{Ablation Study}
To validate the effectiveness of key components, as shown in Tab.~\ref{tab:ablation} and~\ref{tab:ablation 2}, we conduct ablation studies on OpenSatMap20 val set to explore the effectiveness of state update, line reorder, augmentation, the assistance of multi-modal inputs.

\textbf{Effectiveness of State Update.} 
Tab.~\ref{tab:ablation} demonstrates that incorporating state update significantly enhances UniMapGen's performance.
This improvement is attributed to the mechanism that the previous-state map serves as a contextual guide for predicting the current state map, ensuring smooth transitions at patch boundaries. 
Visual comparisons in Fig.~\ref{fig:examples} (b) further reveal that state update generation yields more continuous line structures across patch edges.

\textbf{Effectiveness of Data Augmentation and Reordering Lines.} 
As shown in Tab.~\ref{tab:ablation}, data augmentation significantly enhances model performance, aligning with MLLMs' data-driven characteristics. Line reordering further boosts performance by providing structured sequence patterns, enabling more consistent feature learning and generation.

\textbf{Effectiveness of Multi-modal inputs.} 
Tab.~\ref{tab:ablation 2} reports the model results on the validation subset on nuSences and Argoverse2 cities. Target GT refers to the target GT (a subset of full GT), and Full GT means vanilla full GT.
(1) PV frames. It is observed that both in target and full GT map construction, incorporating PV frames improves the accuracy of constructed maps. This enhancement is due to the complementary ground-level information, which resolves ambiguities in satellite image.
Specifically, PV frames are effective in addressing the occlusion (examples see \textit{suppl.}) and outdateness (Fig.~\ref{fig:examples}(c)) of satellite image.
(2) Text Prompt. 
The BEV-only result on target GT(first line) refers to generating all lane lines without using any target-guided text prompts. Its poor qualitative performance suggests a clear gap between full map generation and target map construction.
With the guidance of text prompts, UniMapGen considerably improved target map construction,
showcasing its ability in target text-prompted generation.
As illustrated in Fig.~\ref{fig:examples} (f), when prompted with road-specific instructions, 
UniMapGen accurately generates specified complete roads, addressing the incomplete annotation of satellite datasets.

\subsection{Qualitative Visualization}

\textbf{PV-based Map Construction.} As shown in Fig.~\ref{fig:examples} (d), when provided with only PV frames, UniMapGen could still generate lane lines that align with the observed lane types, numbers, and location.
This indicates UniMapGen's ability to understand the relationship between PV perspective and lane geometry.
Even in cases where satellite image is outdated, UniMapGen can still produce accurate map representations.

\textbf{Inference Capability.}
UniMapGen demonstrates robust inference ability in challenging scenarios, particularly with significant occlusions from tall structures such as trees (Fig.~\ref{fig:examples} (e)). 
Unlike segmentation-based methods, UniMapGen reconstructs occluded roads even without PV frames by leveraging MLLM-driven in-context reasoning and inference ability rather than relying solely on pixel-level features.

\vspace{-0.5 em}
\subsection{Lane Topology Construction.}
\vspace{-0.3 em}
Due to the absence of intersection annotations in OpenSatMap, we assess UniMapGen's lane topology construction performance on the default nuScenes split.
As shown in Tab.~\ref{tab:topo result}, UniMapGen outperforms leading methods in Reachability F1 and achieves comparable performance in Landmark F1.
Landmark Precision-Recall measures accuracy in locating key points (e.g., splits, forks, merges), while Reachability Precision-Recall evaluates connectivity prediction between them, indicating road existence.
These results show UniMapGen’s effectiveness in constructing complex intersection lane topologies.
Experiment details are in \textit{suppl.}

\section{Limitation and Discussion}
Although UniMapGen demonstrates promising results in large-scale map construction, it has several limitations.
Firstly, the computational demands of MLLMs, along with their relatively slow inference speed (average 4 seconds per 135m$\times$135m), pose practical challenges for real-time map construction like MapTR for autonomous driving.
Actually, UniMapGen serves as a large-scale map constructor to assist autonomous driving and navigation, and could utilize up-to-date PV frames to update existing maps.
Secondly, the text prompt in UniMapGen only supports coordinates for target map construction, but not more flexible natural language, such as generating the map from the school to the hospital. This is due to the lack of training data.
Using an offline object detection model~\cite{carion2020end} to extract the object location or synthesizing the training data leveraging more annotations has the potential to solve this, which will be explored in our future work.

Although UniMapGen has inference ability in map construction, as shown in~\ref{fig:bad case main}, when the lines are extremely worn out or occluded without context (like where the line disappears and reappears among the tall trees), UniMapGen will be confused and construct inaccurate maps.

\section{Conclusion}
We propose a novel generative framework, UniMapGen, for large-scale lane-level map construction. 
UniMapGen formulates lane detection as a discrete sequential generation task, powered by a state update strategy to construct smooth and complete lane lines. Additionally, our framework supports map construction from multi-modal inputs, including BEV images, PV frames, and text prompts, enabling flexible and efficient generation.
In the future, we plan to synthesize more data to enrich our text prompts to support more flexible and user-friendly map construction.
Additionally, we aim to extend our framework to support richer map elements such as traffic signs and road rules.

\bibliography{aaai2026}

\clearpage
\setcounter{page}{1}

\appendix

\twocolumn[
\begin{center}
  {\LARGE \bfseries Supplementary Material to \\ \textit{UniMapGen: A Generative Framework for Large-Scale Map Construction \\ from Multi-modal Data} \par}
  \vspace{1em}
\end{center}
]

This supplementary material provides additional details about the model architecture of UniMapGen, the creation of the training dataset, the data augmentation strategy in training, and the parameters of the models. Additionally, we provide more visualizations of the results of UniMapGen for a better understanding of the ability of our trained model.

\section{Model Architecture}

Each part of UniMapGen in Sec. 3.2 of the main paper is illustrated as follows. During the training process, we fully fine-tune all the parameters in UniMapGen, by updating all the trainable parameters.

\textbf{BEV Encoder.} We employ ViT-Large-14 from Dinov2~\cite{remoteclip} as our pretrained BEV encoder, encoding a single satellite image $I^{BEV}$ into N D-dimensional tokens $\boldsymbol{e}^{BEV} \in \mathbb{R}^{D \times N}$. Dinov2 is trained with a self-supervised learning strategy on a diverse set of 1.2B images, including ImageNet-22k~\cite{deng2009imagenet}, the train split of ImageNet-1k~\cite{deng2009imagenet}, Google Landmarks~\cite{weyand2020google}, and several fine-grained datasets. Please refer to ~\cite{oquab2023dinov2} for more details about Dinov2.

\textbf{PV Encoder.} Ground-level frames provide more detailed, complete, and up-to-date road surface information, complementing satellite image's limitations such as occlusions and outdated data. We utilize a 3D convolution~\cite{tran2015learning} followed by the image encoder from Qwen2-VL-2B~\cite{wang2024qwen2vlenhancingvisionlanguagemodels} (Qwen2-VL-ViT) as our PV Encoder, processing L PV frames $I^{PV}$ into M D-dimensional tokens $\boldsymbol{e}^{PV} \in \mathbb{R}^{D \times M}$.
Qwen2-VL-ViT is a self-implemented ViT by the Qwen team with 675 parameters, which could receive input images at any resolution.

\textbf{Vector Tokenizer and Detokenizer.}
 We propose a special vector tokenizer to convert the previous state map $G_{n-1}$ vector sets into special tokens $\boldsymbol{e}^{G}$, subsequently processed through word embeddings, as mentioned in Sec 3.2 Map Serialization and Model Architecture. 
Specifically, each coordinate number is encoded into special tokens as a whole rather than split into single digits. For example, the number 257 will be tokenized into <257> rather than <2><5><7>. Additionally, the word will be tokenized as a whole as well, such as converting 'points' to <points>. Finally, we remove all the spaces in the token string to reduce the number of tokens.
On the contrary, the vector detokenizer performs the inverse operation, decoding the MLLM-predicted tokens (like the second line) back into vectorized maps (like the first line).

\textbf{Large Language Model and Text Encoder.} 
We adopt Qwen2.5-1.5B~\cite{qwen2025qwen25technicalreport} as our LLM, receiving multimodal input to construct maps. Additionally, we utilize the LLM's text tokenizer to tokenize $T^{prompt}$ into $\boldsymbol{e}^{T}$.
Qwen2.5-1.5B is a dense, decoder-only language model, featuring 1.54 billion parameters and designed for efficient instruction-following tasks. Its architecture incorporates advanced components such as Rotary Position Embeddings (RoPE), SwiGLU activations, RMSNorm normalization, and attention mechanisms with QKV bias and tied word embeddings. The model comprises 28 layers with 12 attention heads for the query and 2 for the key and value, supporting a context length of up to 32,768 tokens and a generation length of 8,192 tokens.
The training process of Qwen2.5-1.5B involved two main stages:
\begin{itemize}
\item Pretraining: The model was trained on a vast corpus encompassing up to 18 trillion tokens, sourced from diverse domains to capture a wide range of knowledge and linguistic patterns

\item Post-training (Instruction Tuning): Following pretraining, the model underwent instruction tuning to enhance its ability to follow user instructions effectively. This stage involved supervised fine-tuning (SFT) using a dataset of over 1 million instruction-output pairs, followed by reinforcement learning from human feedback to align the model's responses with human preferences.

\end{itemize}

\begin{table*}[t]
    \centering
    \renewcommand{\arraystretch}{1.5}
    \setlength{\tabcolsep}{2pt}
    \footnotesize
    \begin{tabular}{l|c}
    \toprule[2pt]
    Modality & Text Prompts \\
    \midrule[1pt]
    BEV only & <image>Please construct the entire road map in the satellite image.  \\
    \midrule[1pt]
    \multirow{2}{*}{PV only} & Please construct the road map referring to the perspective frames: \\
    & [\{<pv frame$_1$>, point: [x$_1$,y$_1$], angle: $\theta_1$\}, ..., \{<pv frame$_L$>, point: [x$_L$,y$_L$], angle: $\theta_L$\}] \\
    \midrule[1pt]
    \multirow{2}{*}{BEV and PV} & <image>Please construct the entire road map in the satellite image, referring to the perspective frames: \\
    & [\{<pv frame$_1$>, point: [x$_1$,y$_1$], angle: $\theta_1$\}, ..., \{<pv frame$_L$>, point: [x$_L$,y$_L$], angle: $\theta_L$\}] \\
    \midrule[1pt]
    BEV and Text (1) & <image>Please construct the road map from (x$_1$,y$_1$) to (x$_2$,y$_2$) in the satellite image.\\
    \midrule[1pt]
    \multirow{2}{*}{BEV and Text (2)} & <image>Please construct the target road map in the satellite image, \\
    & around the trace points: [\{point: [x$_1$,y$_1$], angle: $\theta_1$\},...,\{point: [x$_L$,y$_L$], angle: $\theta_L$\}]\\
    \midrule[1pt]
    \multirow{2}{*}{BEV, PV, and Text} & <image>Please construct the target road map in the satellite image, referring to the perspective frames and trace points: \\
    & [\{<pv frame$_1$>, point: [x$_1$,y$_1$], angle: $\theta_1$\}, ..., \{<pv frame$_L$>, point: [x$_L$,y$_L$], angle: $\theta_L$\}] \\
    \bottomrule[2pt]
    \end{tabular}
    \caption{Text Prompts for each combination of input modalities. <image> refers to the satellite BEV image. <pv frame> refers to the perspective frames. The [x,y] is the PV frame's location in BEV coordinate system, and the angle $\theta$ ranges from 0 to 359.}
    \label{tab:prompt}
\end{table*}

\section{Training Dataset Creation}

As illustrated in Sec. 3.2.2 of the main paper, our UniMapGen supports map construction from multi-modality inputs, including any combination of BEV image, PV frames, and text prompts. 
Tab.~\ref{tab:prompt} shows the text prompts for the map construction from each combination of input modalities.
The details of training data creation for different combinations are depicted as follows.

\textbf{BEV Only Training.} 
We leverage the level-20 parts of OpenSatMap~\cite{zhao2024opensatmap} dataset, denoted as OpenSatMap20, as our BEV images. We use level-20 instead of level-19 because of its higher resolution at 0.15m/pixel.
OpenSatMap20 dataset includes 1180 training and 393 validation satellite image of 4096*4096 pixels. 
Each image in the dataset is annotated with lane-level vectorized maps, with each line classified into three line categories (Curb, Lane line, and Virtual Line), and eight attributes (e.g., line types and level of occlusion).
Following~\cite{zhao2024opensatmap}, we only use line categories (Curb, Lane line, and Virtual Line) and line types (solid line, thick solid line, dashed line, short dashed line and others) as our line attributes of training data.
To meet the training data requirement for MLLMs, we conduct overlapped and inclined cropping data augmentation, creating nearly 700k 896*896 training patches.

\textbf{PV Only Training.}
As illustrated in~\cite{zhao2024opensatmap}, the satellite images in OpenSatMap cover the cities in nuSences and Argoverse2.
We leverage the PV frames from nuSences~\cite{caesar2020nuscenes} and Argoverse2~\cite{wilson2023argoverse} datasets, enabling map construction from PV frames.
To ensure the consistency of training data, we use the modified annotation in OpenSatMap dataset. To be specific, we first use the GPS information provided by nuSences, Argoverse2, and OpenSatMap dataset to link the PV frames to the corresponding BEV images and their annotations. Then, similar to MapTR, we crop the lines within 60m height and 30m width as ground truth, considering the perception field of PV frames.
The PV images are resized into 644*364.

\textbf{PV and BEV Joint Training.} 
Joint training of BEV and PV frames requires paired BEV and PV frames. 
Similarly, GPS information is also used to link PV frames to BEV images between nuSences/Argoverse2 and OpenSatMap dataset. 
The PV position coordinates in the BEV image and the angle representing the vehicle direction are also incorporated into the input to facilitate more accurate alignment between PV and BEV images.
As there may be plenty of PV frames paired with an image patch in OpenSatMap, due to the GPU memory limit,
We only use the front-view frames of these datasets in our experiment.
Additionally, we uniformly sample up to 10 PV images for a paired satellite image.
The PV images are resized into 644*364 resolution.

\textbf{ Text-prompted Target Map Generation. }
We use a semi-automatic annotation pipeline: automatic data synthesis via Segment Anything Model (SAM), followed by human verification and correction for accurate target lane lines and text prompts matching.
Additionally, we also incorporate the trace points from nuSences and Argoverse2 as text prompt for target map generation.
To be specific, the semi-automatic annotation includes automatic processing to synthesize the training data and human-assisted post-processing to check and correct the inaccurate synthesized data.
\begin{itemize}
\item \textbf{Automatic processing}. This aims to create the training data of text prompts for the targeted map paired with the corresponding targeted map vectors.
To achieve this, we utilize the Segment Anything Model (SAM)~\cite{kirillov2023segment} to generate precise road area segmentations for each line vector annotation in OpenSatMap. Specifically, we input the annotated points from line vectors directly into SAM, which effectively segments the corresponding road regions.
To identify complete road segments and their associated line vectors, we employ a region merging algorithm. Specifically, we merge segmented regions with an Intersection over Union (IoU) threshold of 0.5, while simultaneously combining their corresponding line vector annotations. This process results in comprehensive road segments paired with their complete set of line vector representations.

\item \textbf{Human-assisted post-processing.}
We employ three expert annotators to check and correct all the inaccurate automatically synthesized segmentation-line pairs.
Finally, we automatically extract the start and end points of each segmentation road, creating the start/end points-line pairs, which are the training data for targeted map generation.

\end{itemize}

\section{Lane Topology Construction}
As mentioned in Sec. 4.4 in main paper, due to the absence of intersection annotations in OpenSatMap, we assess UniMapGen's lane topology construction performance on the default nuScenes split.
The model is trained for 30 epochs with a batch size of 32. The learning rate is set to 1e-4 with cosine learning rate decay adopted.
We use Landmark/Reachability Precision-Recall to evaluate the model's performance on lane topology construction.
Landmark Precision-Recall measures accuracy in locating key points (e.g., splits, forks, merges), while Reachability Precision-Recall evaluates connectivity prediction between them, indicating road existence.
Together, these metrics offer a comprehensive evaluation of the lane graph.

\section{Parameters}

\textbf{Training Parameters.} We trained our UniMapGen for 6 epochs with a batch size of 32 and the fully fine-tuning strategy applied. We used BF16 mixed precision training with DeepSpeed~\cite{rasley2020deepspeed} to increase the training efficiency.
We used the AdamW~\cite{loshchilov2017decoupled} optimizer and a weight decay of 0.1. A cosine learning rate decay with a peak learning rate of 2e-5 and a linear warm-up of 100 steps was adopted.

\textbf{Model Parameters.} Tab.~\ref{tab:para vit} illustrates the model parameters of BEV encoder (Dinov2-Large) and PV encoder (Qwen2-VL-ViT). Tab.~\ref{tab:para llm} shows the model parameters of the large language model Qwen2.5-1.5B.

\begin{figure*}[t]
    \centering
    \includegraphics[width=1.0\textwidth]{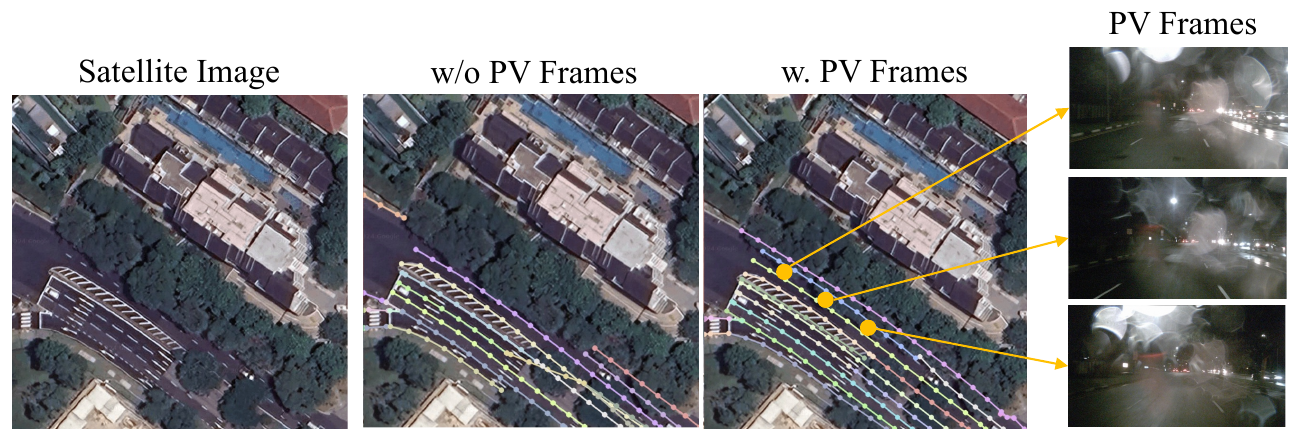}
    \caption{Examples of PV and BEV joint map construction. With the help of PV frames to support clearer road information, UniMagGen can construct the maps occluded by tall trees. Please zoom in for better visualization.
    }
    \label{fig:pv bev}
\end{figure*}

\begin{table*}[!htbp]
    
    \centering
    \begin{tabular}{l|ccccc}
        \toprule[2pt]
        Model & Layers & Hidden size D & MLP size & Heads & Params  \\
        \midrule[1pt]
        Dinov2-Large  & 24 & 1024 & 4096 & 16 & 307M \\
        Qwen2-VL-ViT  & 32 & 1536  & 6144 & 16 & 675M \\
        \bottomrule[2pt]
    \end{tabular}
    \caption{Model parameters of image encoder model (Vision Transformer).}
    \label{tab:para vit}
\end{table*}

\begin{table*}[!htbp]
\centering

\begin{tabular}{lccccc}
\toprule[2pt]
Models & Layers & Heads (Q / KV) & Tie Embedding & Context / Generation Length \\
\midrule[1pt]

Qwen2.5-1.5B   & 28     & 12 / 2         & Yes           & 32K / 8K                    \\
\bottomrule[2pt]
\end{tabular}
\caption{Model parameters of Qwen2.5-1.5B.}
\label{tab:para llm}
\end{table*}

\section{Visualizations}
\textbf{PV and BEV joint map construction.} 
Fig.~\ref{fig:pv bev} shows an example of map construction with/without PV frames. The sample comes from Singapore hollandvillage, and the PV frames are selected from the nuSences datasets.
It is observed that  PV frames support the clear road information of the regions occluded by tall trees. Therefore, with the help of PV frames, UniMapGen can accurately construct the maps occluded by the trees.

\begin{figure*}[t]
    \centering
    \includegraphics[width=1.0\textwidth]{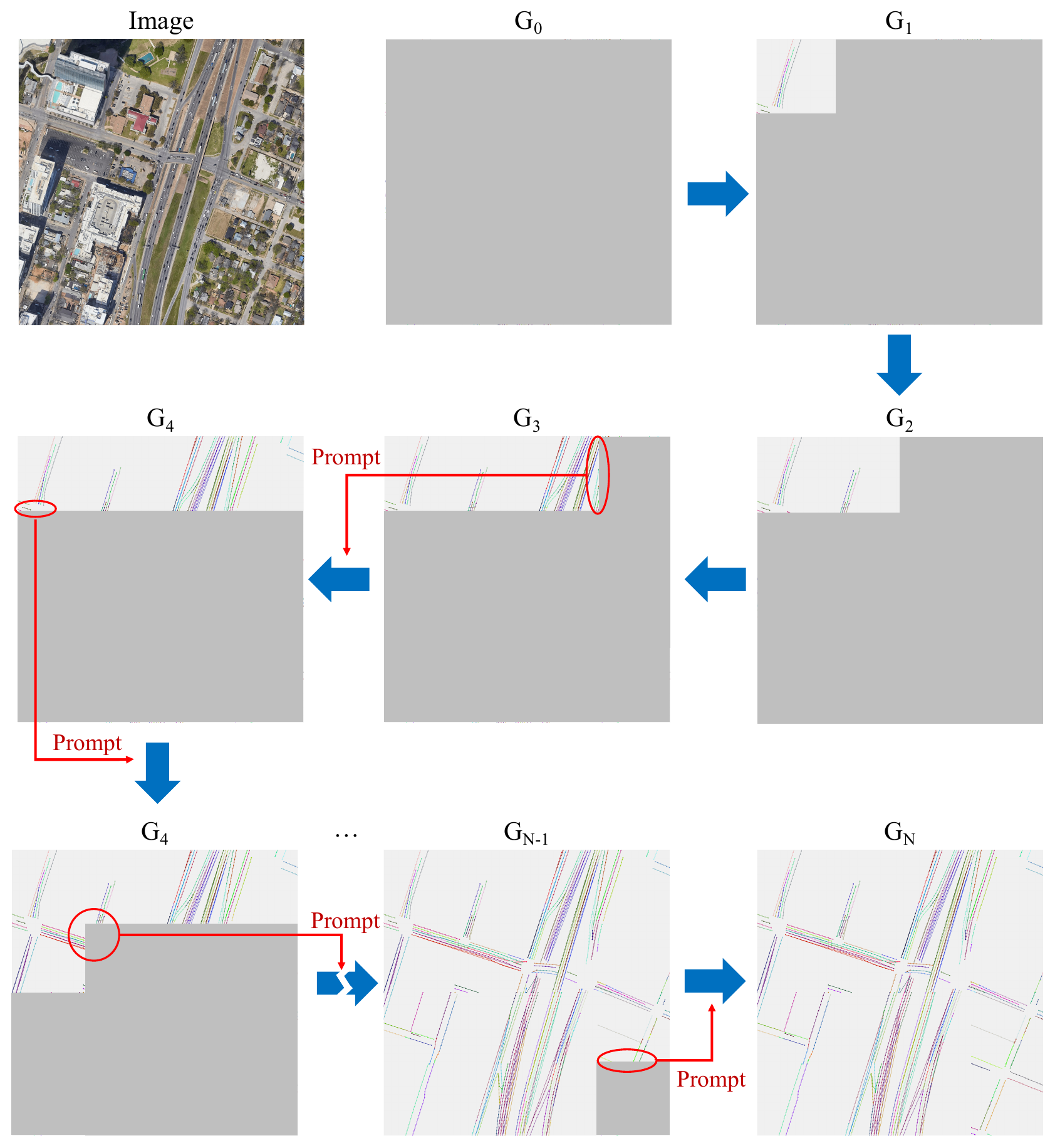}
    \caption{Examples of the process of large-scale global map construction by state update. The current-state map is constructed prompted by the previous-state map.
    }
    \label{fig:state updating}
\end{figure*}

\textbf{State Update.} 
Fig.~\ref{fig:state updating} shows an example of large-scale global map construction by state updating. 
For practical implementation in the OpenSatMap experiments, we adopted a left-to-right, top-to-bottom region update strategy. 
The state update process is similar to a growth process, where the current map evolves from the previous map.
The previous map acts as a prompt for the map construction of the current state to generate global continuous and consistent map.

\begin{figure*}[!htbp]
    \centering
    \includegraphics[width=0.8\textwidth]{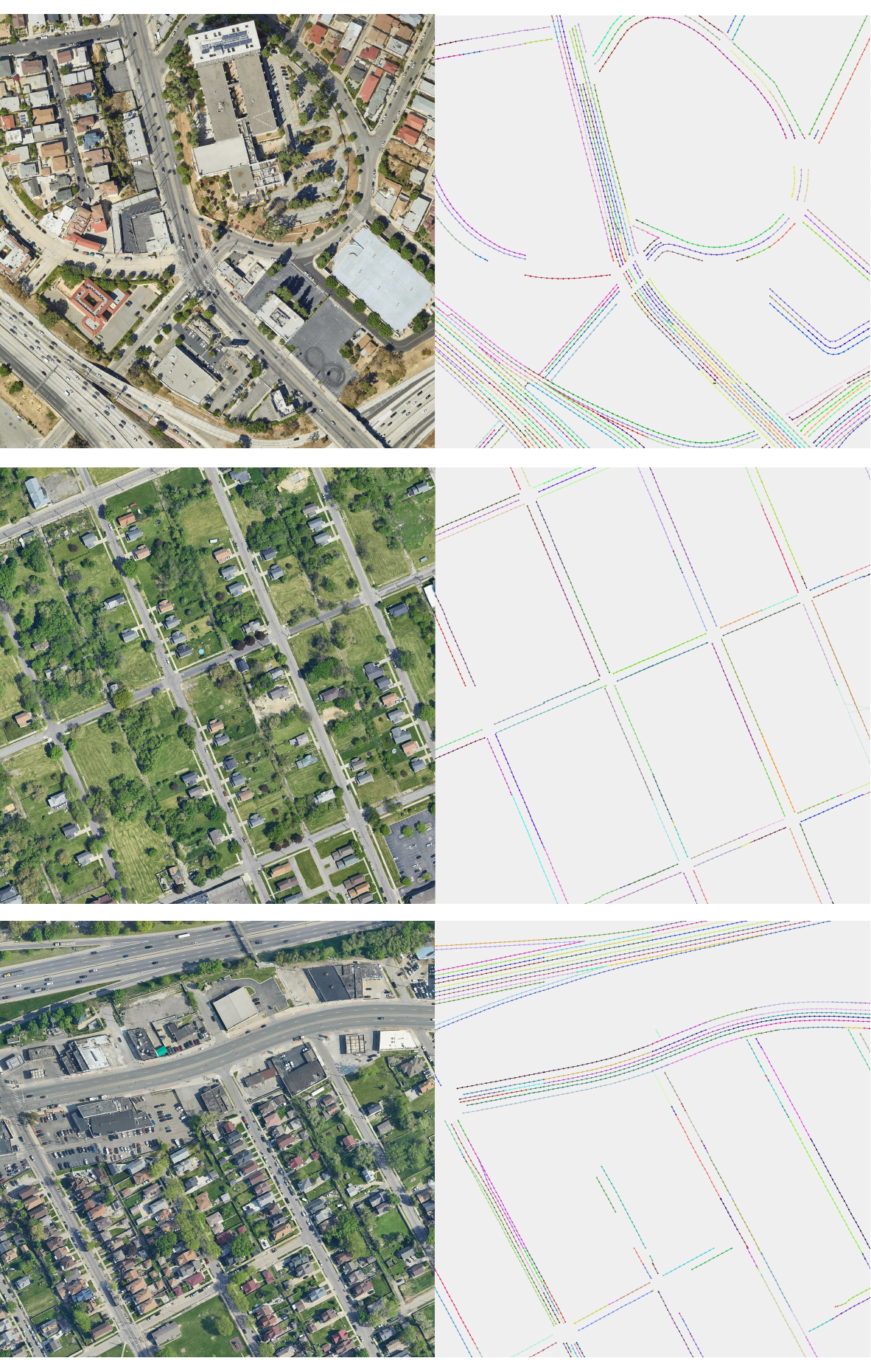}
    \caption{Examples of large-scale map construction by UniMapGen. (zoom in for better look) 
    }
    \label{fig:all example 1}
\end{figure*}

\begin{figure*}[!htbp]
    \centering
    \includegraphics[width=0.8\textwidth]{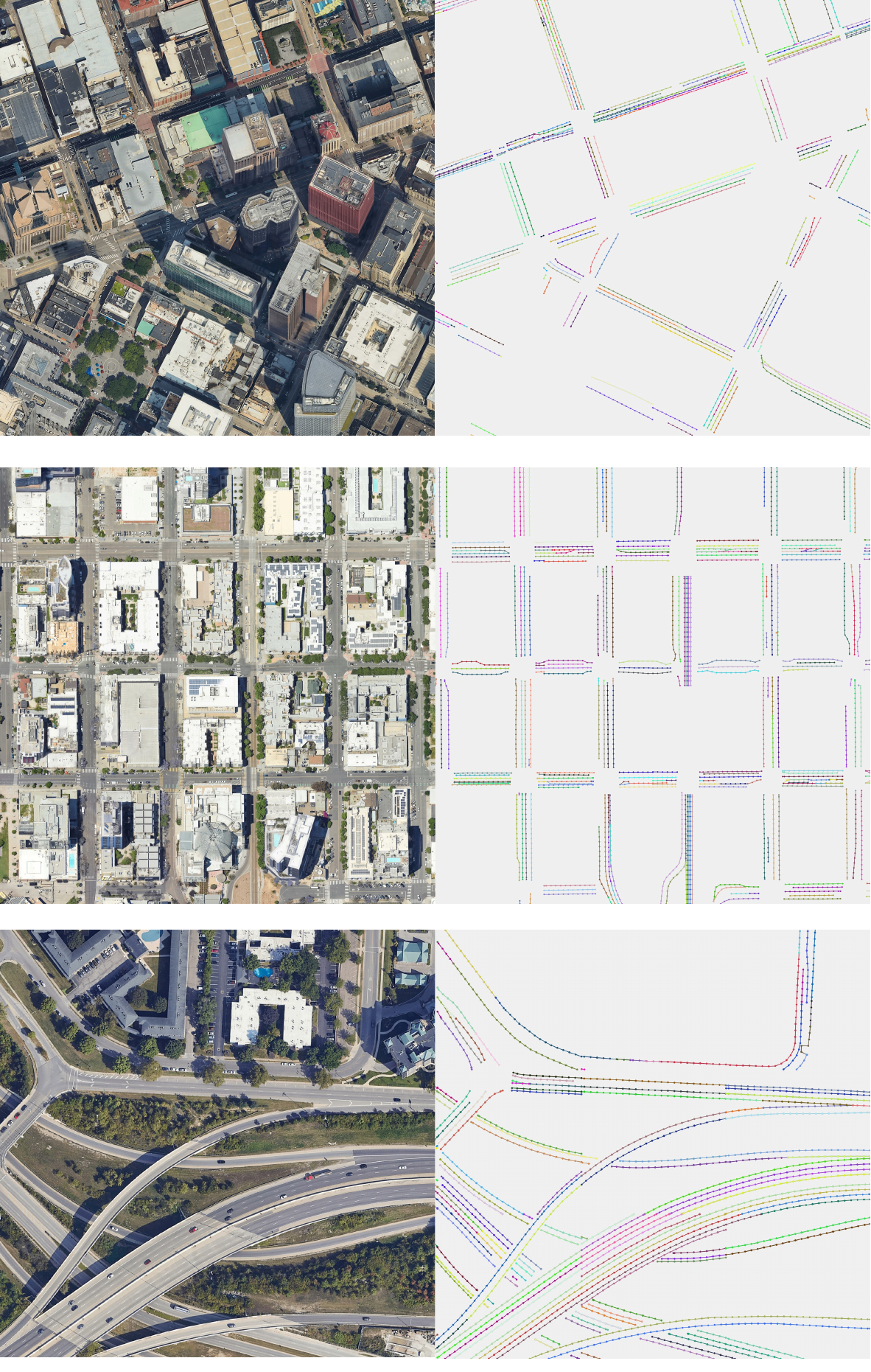}
    \caption{Examples of large-scale map construction by UniMapGen. (zoom in for better look) 
    }
    \label{fig:all example 2}
\end{figure*}

\textbf{Large-Scale Map Construction.}
Fig.~\ref{fig:all example 1} and~\ref{fig:all example 2} illustrate several examples of large-scale map construction by UniMapGen.
UniMapGen constructs large-scale global continuous and consistent maps in an end-to-end manner without post-processing process.

\end{document}